\documentclass[journal,twoside,web]{ieeecolor}
\usepackage{tmi}
\usepackage{cite}
\usepackage{amsmath,amssymb,amsfonts}
\usepackage{algorithmic}
\usepackage{graphicx}
\usepackage{textcomp}
\usepackage[ruled,vlined]{algorithm2e}
\usepackage{hyperref}
\usepackage{bm}

\usepackage{subcaption}
\usepackage{array}
\usepackage{booktabs}
\usepackage{caption}  %
\usepackage[table,xcdraw]{xcolor}
\usepackage{wrapfig}
\usepackage{verbatim}
\usepackage{color}

\def\BibTeX{{\rm B\kern-.05em{\sc i\kern-.025em b}\kern-.08em
    T\kern-.1667em\lower.7ex\hbox{E}\kern-.125emX}}
\begin{document}
\title{Data-iterative Optimization Score Model for Stable Ultra-Sparse-View CT Reconstruction}
\author{ Weiwen Wu, Yanyang Wang 
\thanks{W. Wu, Y. Wang are with the Department of Biomedical Engineering, Sun-Yat-sen University, Shenzhen Campus, Shenzhen, China. The contribution of W. Wu and Y. Wang are equal. Wu is the corresponding author (e-mail: wuweiw7@mail.sysu.edu.cn).}% <-this % stops a space
}

\maketitle

\begin{abstract}
Score-based generative models (SGMs) have gained prominence in sparse-view CT reconstruction for their precise sampling of complex distributions. In SGM-based reconstruction, data consistency in the score-based diffusion model ensures close adherence of generated samples to observed data distribution, crucial for improving image quality. Shortcomings in data consistency characterization manifest in three aspects. Firstly, data from the optimization process can lead to artifacts in reconstructed images. Secondly, it often neglects that the generation model and original data constraints are independently completed, fragmenting unity. Thirdly, it predominantly focuses on constraining intermediate results in the inverse sampling process, rather than ideal real images. Thus, we propose an iterative optimization data scoring model. This paper introduces the data-iterative optimization score-based model (DOSM), integrating innovative data consistency into the Stochastic Differential Equation, a valuable constraint for ultra-sparse-view CT reconstruction. The novelty of this data consistency element lies in its sole reliance on original measurement data to confine generation outcomes, effectively balancing measurement data and generative model constraints. Additionally, we pioneer an inference strategy that traces back from current iteration results to ideal truth, enhancing reconstruction stability. We leverage conventional iteration techniques to optimize DOSM updates. Quantitative and qualitative results from 23 views of numerical and clinical cardiac datasets demonstrate DOSM's superiority over other methods. Remarkably, even with 10 views, our method achieves excellent performance.
\end{abstract}


\section{Introduction}
Computed Tomography (CT) has garnered extensive utilization in both medical diagnosis due to its ability to provide practical and precise diagnostic outcomes \cite{kazerouni2022diffusion}. Sparse-view CT scanning emerges as a promising strategy to decrease radiation dose, wherein only a fraction of projection data is required for image reconstruction\cite{willemink2019evolution} \cite{han2018framing}. However, the limited number of measurement views leads to a reduced acquisition \cite{bian2010evaluation} of prior information within imaged object, consequently yielding a deterioration in the image quality  \cite{wu2021drone}. Moreover, the diminished rank of the measurement matrix introduces an augmented array of potential solutions\cite{niu2018nonlocal}, impeding convergence and culminating in a heightened prevalence of uncertainties and inaccuracies \cite{ben2002robust} \cite{raj2020improving}.

Traditional reconstruction methods like filtered back-projection (FBP) yield unsatisfactory results, marked by streaking artifacts and poor image quality \cite{willemink2019evolution}. The advent of artificial intelligence has fostered the creation of advanced deep-learning techniques for enhancing sparse-view CT reconstruction. Notable examples of such techniques encompass FBP-ConvNet \cite{jin2017deep}, Densenet Deconvolution Network \cite{zhang2018sparse}, residual encoder-decoder convolutional neural network \cite{chen2017low}, generative adversarial network (CGAN) \cite{ghani2018deep}, and multi-domain integrative swin transformer \cite{pan2022multi}, among others. These deep-learning approaches primarily hinge on supervised training with paired images to yield superior outcomes.

Score-based generative models (SGMs) have risen to prominence due to their exceptional capability to precisely sample from intricate distributions \cite{ho2020denoising}  \cite{songscore}. For example, denoising diffusion probabilistic models (DDPM)\cite{ho2020denoising}, denoising diffusion restoration models (DDRM) \cite{kawar2022denoising}, Denoising diffusion implicit models (DDIM) \cite{song2020denoising}, Stochastic Differential Equations (SDE) \cite{songscore}, etc. Typical SGMs follow a two-stage process consisting of a forward stage perturbing data to noise and a reverse stage converting noise back to data. The reverse process is usually achieved by parameterized deep neural networks, e.g., an SGM, which can be optimized with the training data. The power of SGM seamlessly extends to various applications within the field of medical image reconstruction \cite{songsolving} \cite{li2023descod}, particularly in scenarios such as sparse-view CT reconstruction with remarkable achievements \cite{xia2022patch} \cite{li2023two}. Data consistency in a score-based score-based model is to ensure that the generated samples or images adhere closely to the observed data distribution, which plays an important role for improving reconstructed image quality \cite{chung2022improving}. Data consistency acts as a constraint during the sampling process to align the generated samples with the actual data, improving the quality and realism of the generated samples \cite{chung2022diffusion}. For a typical score-based generative model, the data consistency is enforced \cite{karras2022elucidating} by iteratively adjusting the generated samples based on the gradient of the log-likelihood of the observed data \cite{songscore}. This process aims to reduce the discrepancy between the generated samples and the real measurement data, resulting in better alignment and more accurate modeling of the underlying data distribution \cite{kazerouni2023diffusion}. By maintaining data consistency \cite{chung2022come}, the model becomes more capable of generating samples that are representative of the original data.

These existing SGMs for sparse-view CT reconstruction mainly focus on performing the diffusion within sinogram-based and image-domain. In respect to the sinogram-based SGM, the data consistency is formatted by utilizing the generated full sinogram and the sampling location of the original sparse-view data. For instance, a notable advancement involves the integration of a fully unsupervised score-based generative model into the sinogram domain, effectively enhancing sparse-view CT reconstruction capabilities \cite{guan2022generative}. Furthermore, a patch-based denoising diffusion probabilistic model was tailored and developed \cite{xia2022patch}. In the image-domain context, the conventional data consistency hinges on a sparse-sampling mask \cite{songsolving}. The data consistency process involves employing partial projections derived from the reconstructed image, wherein $\bm{y}_t=\bm{Ax}_t$, signifying the system matrix $\bm{A}$ and reconstructed image $\bm{x}_t$ at time step $t$. However, owing to the inherent imperfections within the interim $\bm{y}_t$, inaccuracies manifest in the estimated data $\bm{y_t^{'}}=mask*\bm{y}+(1-mask)*\bm{y}_t$ ($\bm{y}$ is the sparse-view data), which in turn contributes to data consistency discrepancies, subsequently introducing secondary artifacts in the reconstruction process. Moreover, past investigations frequently updated the data consistency and generative model as separate entities, disregarding the intrinsic link between these modules \cite{chung2022improving}\cite{chung2022diffusion}. These approaches not only hinder the model's ability to converge to a stable reconstruction outcome but can also exacerbate the inherent instability within the reconstruction process.

\begin{figure}[t]
    \centering
    \vspace{-0.cm}
    \includegraphics[scale=0.4]{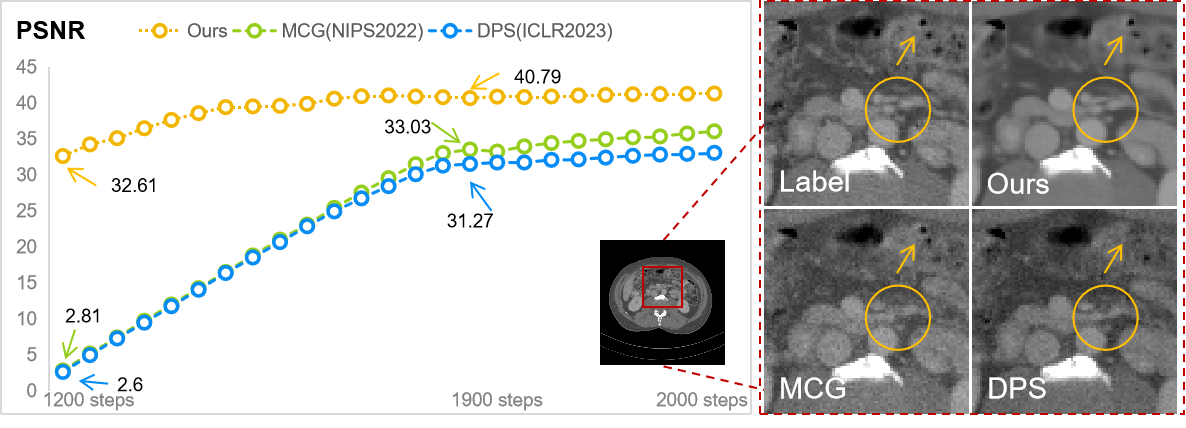}
    \caption{A comprehensive evaluation contrasting our proposed methodology with contemporary state-of-the-art reconstruction techniques using 23 views' measurements.}
    \label{motion}
\end{figure} 

Overcoming the lack of precise data consistency constraints is a significant challenge when striving for high-quality image reconstructions, especially in the context of ultra-sparse-view CT reconstruction \cite{wu2021drone}. In this scenario, the sparse-view sampling mask can introduce projection errors, leading to the presence of erroneous artifacts \cite{sidky2006accurate}. To mitigate inaccuracies in projection, a more effective approach involves utilizing measurement data solely for establishing data consistency, defined as $\|\bm{y}-\bm{Ax}\|_2^2$. This approach ensures the generation of reliable reconstruction outcomes. Importantly, the focus of data consistency should be directed towards $\bm{x}_0$ rather than $\bm{x}_t$ at time $t$. In simpler terms, when $\bm{y}\ne \bm{Ax}_t$, it becomes insufficient to drive the generation process towards satisfactory results. A viable solution is to enforce data consistency as $\|\bm{y}-\bm{Ax}_0 (\bm{x}_t )\|_2^2$. However, a significant challenge arises in establishing an explicit relationship between $\bm{x}_t$ and $\bm{x}_0$. This is due to the fact that $p(\bm{x}_0)$ follows a posterior distribution, which remains unknown during the inference stage \cite{chung2022improving} \cite{chung2022diffusion}. Furthermore, it is imperative to devise a viable approach for augmenting the coherence between the data consistency and generative update. This strategy's purpose should encompass the simultaneous constraint of both data consistency and generative outcomes, effectively capturing the intricate correlations inherent to these two modules. To fully harness the potential of data consistency within the score-based model for tackling inverse problems, pursuing a strategy that explores this avenue comprehensively is both valuable and promising.

In this paper, we introduce an innovative data consistency component to the SDE, which serves as a valuable constraint in the context of ultra-sparse-view CT reconstruction. We term this framework the Data-iterative Optimization Score-based Model (DOSM). The distinctive aspect of this new data consistency element lies in its utilization of solely the original measurement data to confine the outcomes of generation, effectively sidestepping any undesirable structures. Furthermore, we establish a pioneering inference strategy that traces back from $\bm{x}_t$ to $\bm{\hat{x}}_0$ using a solid theoretical foundation at any time $t$. In contrast to prevailing state-of-the-art techniques, our approach exhibits a noteworthy enhancement in both image quality and quantitative outcomes (PSNR and SSIM), as indicated in Fig. \ref{motion}. The significant contributions made by this study can be succinctly outlined as follows:

\begin{itemize}
\item[1)]
We ingeniously devise a novel data consistency term that harmoniously balances the constraints from measurement data and the generative model. This seamless integration facilitates the capture of intricate correlations and significantly enhances their alignment. By fusing this innovative data consistency approach with an SDE, we introduce an entirely new score-based model for ultra-sparse-view CT reconstruction.
\end{itemize}

\begin{itemize}
\item[2)]
We develop an innovative and adaptable inference strategy that elegantly traces from $\bm{x}_t$ back to its initial counterpart $\bm{\hat{x}}_0$. Remarkably, as the number of iterations decreases, the accuracy of the estimated $\bm{\hat{x}}_0$ improves significantly. The estimated $\bm{\hat{x}}_0$ is introduced into the formulated data consistency for formulating a novel unified reconstruction model.
\end{itemize}

\begin{itemize}
\item[3)]
We employ a suitable solving algorithm to optimize the established score-based Model. To ensure solution stability, we incorporate the wisdom of the conventional iteration reconstruction technique to optimize the update of data consistency. 
\end{itemize}

\begin{itemize}
\item[4)]
The efficacy and advantages of our proposed approach are meticulously examined and evaluated. We validate our method using extensive sparse-view CT datasets, encompassing both numerical simulations and real clinical cardiac data, showcasing its superior reconstruction performance in comparison to state-of-the-art alternatives.
\end{itemize}

\section{Related Works}

\subsection{ Stochastic Differential Equation Models}
The fractional derivative-based generative model, known as SDE, stands as a quintessential example within the realm of generative models. It has a widespread utilization application in the domain of medical imaging for addressing inverse problems. SDE gradually inject the noise onto the clean data, shaping the training process of neural networks, which is termed as the forward process. Subsequently, it delineates a perturbation process that inverses the initial forward step, progressively denoising the noisy samples. This iterative denoising process is referred to as the inverse process, embodying the image reconstruction phase in image reconstruction. Notably, within this framework, Y. Song et al\cite{songscore} introduced a fractional score-based Model to define the forward process. Throughout the entirety of the diffusion process, data $\bm{x}$ is represented as $\bm{x}(t)=\bm{x}_t$, with $t\in [0,1]$. Here, $\bm{x}_0$ signifies the distribution of the original training data, while $\bm{x}_T$ approximates a spherical Gaussian distribution and $T$ represents the number of diffusion step in forward process. The formulation of the forward process is articulated as follows \cite{songscore}:
\begin{equation}\label{eq1}
d\bm{x} = f(\bm{x},t)dt + g(t)d\bm{w},
\end{equation}
where $f$ delineates the linear drift function, and $g(t)$ characterizes the scalar diffusion coefficient. The vector $\bm{w}$ follows the standard Brownian motion. The inverse process of Eq. \eqref{eq1} can be formulated as follows:
\begin{equation}\label{eq2}
\begin{aligned}
d\bm{x} &= \left[ {f(\bm{x},t) - g{{(t)}^2}{\nabla _{\bm{x_t}}}\log {p_t}(\bm{x})} \right]dt + g(t)d\bar{\bm{w}},
\end{aligned}
\end{equation}
where $dt$ represents an infinitesimal time step. In order to generate reconstructed images through the inverse process of SDE, a time-dependent score function ${\nabla {\bm{x_t}}}\log {p_t}(\bm{x})$ is required \cite{songsolving}. This scoring function estimator ${\bm{s_\theta}}$ can be obtained by means of denoising scoring matching \cite{song2019generative}:
\begin{eqnarray}\label{eq3}
\begin{aligned}
\min _{\boldsymbol{\theta}} \mathbb{E}_{\boldsymbol{x}_{t} \mid \boldsymbol{x}_{0}, \boldsymbol{x}_{0}}\left[\left\|\boldsymbol{s}_{\boldsymbol{\theta}}(\bm{x}_t, t)-\nabla_{\bm{x}_{t}} \log p\left(\bm{x}_{t} \mid \bm{x}_{0}\right)\right\|_{2}^{2}\right].
\end{aligned}
\end{eqnarray}
We estimate the score function $\theta^*$ through training a neural network. The fractional scoring estimator ${\bm{s_\theta}}(\bm{x}_t,t)$ can be employed as a replacement for the scoring function in Eq. \eqref{eq2}. Different combinations of $f$ and $g$ can generate different types of SDE. For the typical variation exploding (VE) -SDE, $f$ and $g$ are set as follows \cite{songscore}:
\begin{equation}\label{eq4}
f = 0,~~~  g = \sqrt {\frac{{d\left[ {{\sigma ^2}(t)} \right]}}{{dt}}},
\end{equation}
where $\sigma{(t)}$ signifies the time-varying escalating scale function for noise.

\subsection{Score-based Sparse-View CT Reconstruction}
Score-based Model shows good performance in sparse-view CT reconstruction \cite{kazerouni2022diffusion}. Relying on the powerful generation ability of the score-based model, there are some typical works including \cite{guan2022generative}, \cite{chung2022improving} and etc. The sparse-view reconstruction can be regarded as the classical inverse problem \cite{songsolving}, which aims at recovering an unknown signal from a set of observed signals. Specifically, if $\bm{x}$ is an original signal and $\bm{y = Ax+\epsilon}$ is a noisy observation given by a set of linear measurements. The sampling procedure $\bm{A}$ is the radon transform. The sparse-view reconstruction is equivalent to recovering the signal $\bm{x}$ from the measurement $\bm{y}$.  The $\bm{\varepsilon}$ denotes the measured noise. A common practice is to use the least squares of linear models to solve image reconstruction \cite{zibetti2018total}. Due to the uncertainty of the inverse problem (where $\bm{A}^{T}\bm{A}$ is singular or ill-conditioned, and the data is noisy), the standard practice is to introduce regularization terms for reconstruction:
\begin{eqnarray}\label{eq5}
\bm{x} = \underset{\bm{x}}{\operatorname{argmin}} \frac{1}{2}\|\boldsymbol{y}-\boldsymbol{A} \boldsymbol{x}\|_{2}^{2}+\lambda R(\boldsymbol{x}),
\end{eqnarray}
where $R(.)$ is a suitable regularization and it can be a sparse transformation regularization.

%The solution of Eq. \eqref{eq1} satisfies:
%\begin{eqnarray}\label{eq2}
%\bm{A}^{T} \bm{A} \bm{x}_{x}+\lambda \nabla R\left(\bm{x}\right) & = & \bm{A}^{T} %\bm{y}.
%\end{eqnarray}
\subsection{Data Consistency Construction}
The good reconstruction performance of the diffusion model not only depends on the generation of the unconditional fractional function ${\nabla _{\bm{x_t}}}\log {p_t}(\bm{x})$, but also needs a suitable data consistency strategy to guide the generation process. In solving CT imaging problems, the data consistency strategy was first proposed by Y. Song et al \cite{songsolving}. Specifically, the observed projection information, $y$, and the sinusoidal graph generated at each step are completed. 
\begin{eqnarray}\label{eq7}
\bm{x}_t^{\prime}= \bm{A}^T[\bm{y}*mask+A\bm{x}_t*(1-mask)],
\end{eqnarray}
where $\bm{x}_t^{\prime}$ is the result of adding measurement data constraints. $mask$ represents to extract sparse-view data from the full-view projection.

The other is a data consistency strategy based on posterior probability sampling, which provides an easy-to-handle approximation for $p(\bm{y|x_t})$ in order to better solve the reconstructed inverse problem using the scoring function \cite{chung2022diffusion}. 
\begin{eqnarray}\label{eq8}
\nabla_{\boldsymbol{x}_{t}} \log p\left(\boldsymbol{y} \mid \boldsymbol{x}_{t}\right) \simeq-\frac{1}{\sigma^{2}} \nabla_{\boldsymbol{x}_{t}}\left\|\boldsymbol{y}-{A}\left(\hat{\boldsymbol{x}}_{0}\left(\boldsymbol{x}_{t}\right)\right)\right\|_{2}^{2},
\end{eqnarray}
where $\sigma$ is the measurement noise factor. $\hat{\boldsymbol{x}}_{0}$ is the posterior mean calculated by the diffusion model and is a function of $x_t$.

To further improve the constraint generation process, Chung  et al. proposed a correction term for manifold constraints \cite{chung2022improving}. Hence, a constraint based on manifold gradients is formulated to bound the gradient of the measurement term within the confines of the data manifold.
\begin{eqnarray}\label{eq9}
\nabla_{\boldsymbol{x}_{t}} \log p\left(\boldsymbol{y} \mid \boldsymbol{x}_{t}\right) \simeq-\alpha \frac{\partial}{\partial \boldsymbol{x}_{t}}\left\|\boldsymbol{W}\left(\boldsymbol{y}-\boldsymbol{A} \hat{\boldsymbol{x}}_{0}\left(\boldsymbol{x}_{t}\right)\right)\right\|_{2}^{2},
\end{eqnarray}
where $\alpha$ corresponds to the step length, and $\boldsymbol{W}$ denotes a weighting factor. The authors in \cite{chung2022improving} additionally make use of the projected estimate obtained from intermediate updates.

The first and third data consistency terms, due to inherent imperfections in estimated data, introduce secondary artifacts during reconstruction. The second data consistency term prioritizes derivative descent, making it sensitive to perturbations and destabilizing the model. Additionally, these data consistency strategies consider multiple task mappings but overlook CT-specific physical priors. Simple FBP reconstruction limits the generative model's exploration of its potential. Importantly, these three data consistency terms frequently update data and the generative model separately, ignoring their intrinsic connection. This not only obstructs stable convergence but can worsen inherent reconstruction instability.

\section{Methods}
%%%
\subsection{Mathematical Model}
Take into consideration a broad-ranging forward model for computed tomography reconstruction, the ideal reconstruction solution of $\bm{{x_0}}$ should satisfy:
\begin{equation}\label{eqIII1}
\bm{A}\bm{{x_0}}+\bm{\varepsilon}=\bm{y},
\end{equation}
%The symbols $\bm{A}$ and $\bm{y}$ represent the system matrix after projection and the captured measurement data, respectively. 
Let's remember that $\bm{A}$ embodies the scanner's operation, referred to as forward projection. Each individual row within $\bm{A}$ encapsulates the coefficients of an equation that correlates with a singular ray. It describes how the pixels are combined into ray sums. Here, $h_{ij}$ denotes an element located at the $(i,j)-th$ position, which $i$ takes values from 1 to $I$, and $j$ spans from 1 to $J$. $I$ corresponds to the total pixel count within the reconstructed image, while $J$ is the number of X-ray pathways. 
In the score-based model solving the CT reconstruction problem, the approximate fractional function ${\bm{s_\theta}}$ obtained from the forward process training can simulate the reverse denoising process in the reconstruction process \cite{songsolving}. ${\bm{s_\theta}}$ is approximated as ${\nabla _{\bm{x_t}}}\log {p_t}(\bm{x})$. Since there is no explicit correspondence \cite{chung2022diffusion} between $\bm{y}$ and $\bm{x}$, we cannot solve it directly using the neural network of the score-based Model. A common practice \cite{chung2023solving} is to solve the posterior probability according to Bayesian theory $P(\bm{x|y})=p(\bm{x})p(\bm{y|x})/p(\bm{y})$.
Specifically, the reconstructed image $\bm{x}_t$ at the current time can be regarded as the inverse solution through the observation data $\bm{y}$:
\begin{equation}\label{eqIII2}
\nabla_{\bm{x}_t}log\left(p\left(\bm{x}_t|\bm{y}\right)\right)=\nabla_{x_t}log\left(p\left(\bm{x}_t\right)\right)+\nabla_{\bm{x}_t}log\left(p\left(\bm{y}/\bm{x}_t\right)\right).
\end{equation}
Considering that $y$ and $x_t$, $x_0$ have no explicit correspondence, they cannot be solved directly by Eq. \eqref{eqIII2}. So we construct an approximate relationship:
\begin{equation}\label{eqIII3}
\nabla_{\bm{x}_t}log\left(p\left(\bm{y}|\bm{x}_t\right)\right)\approx\nabla_{\bm{x}_t}log\left(p\left(\bm{y}|\hat{\bm{x}_0}\left(\bm{x}_t\right)\right)\right).
\end{equation}
Utilizing the diffusion model for addressing sparse-view CT reconstruction, the mathematical formulation of the model can be expressed as follows:
\begin{equation}\label{eqIII4}
\min_{\bm{x}_t} M(\bm{x}_t) , ~~s.t., \left \| \bm{A}\hat{\bm{x}_0}(\bm{x}_t)-\bm{y}  \right \| _{2}^{2}\le {\tau}  
\end{equation}
Here, the term $M\left(\bm{x}_t\right)$ signifies the solver configured through Stochastic Differential Equations (SDE) at any time t. 
%The SDE can be refined by integrating a standard Predictive Corrector (PC) approach. The generation process of the diffusion model involves a comprehensive numerical solution to the inverse SDE outlined in Eq. \eqref{eqIII3}. To achieve this, we employ a classical predictive correction method that guides solving the inverse SDE.
Furthermore, Eq. \eqref{eqIII4} encompasses the solution for a condition-constrained diffusion model. In essence, it can be transformed into an unconstrained optimization problem, characterized by the subsequent expression:
\begin{equation}\label{eqIII5}
\bm{x}_{t-1}=\min_{\bm{x}_t}\frac{1}{2} \left \| \bm{A}\hat{\bm{x}_0}(\bm{x}_t)-\bm{y}  \right \| _{2}^{2}+ \eta_t M(\bm{x}_t)
\end{equation}
where $\eta_t$ embodies the equilibrium coefficient that finely tunes the accentuation of the data consistency proportion at the temporal instant denoted as $t$. 

%\subsection{${\bm{x}_0}$'s Estimation}
As introduced above,  the posterior probability sampling of the score-based model depends on $\boldsymbol{{x}}_0$, i.e., the underlying ground truth reconstructed image. In this case, effective estimation of $\bm{{x}}_0$ becomes a critical step to conduct the posterior sampling. The estimation is usually performed with the latest intermediate result of $\bm{x}_{t}$. For example,  $\boldsymbol{\hat{x}}_0 (\boldsymbol{x}_{t}) = \bm{x}_{t} + \sigma _{t}^{2} \nabla_{\boldsymbol{x}_{t}} \log p\left(\boldsymbol{x}_{t}\right)$ is applied in \cite{lee2023improving, chung2022improving} to achieve
the denoised result by computing the posterior expectation according to Tweedie's formula \cite{efron2011tweedie}. We argue that although this estimation enjoys simplicity, it's sole dependency on $\boldsymbol{x}_{t}$ may probably lead to an unstable and biased estimation, resulting in a sub-optimal solution to the posterior sampling. To obtain a more reliable estimation, this paper innovatively proposes an effective method with $\bm{x}_{t}$s from multi-channels, as shown in Fig. \ref{mainfig1}. Specifically, once the diffusion model is trained, the generation process will be conducted in multiple, say $N$, channels starting from independently sampled noise respectively. At each intermediate step, $\bm{{x}}_0$ can be estimated by a weighted combination of the intermediate results from the $N$ channels as follows:
\begin{equation}\label{eqIII6}
\boldsymbol{\hat{x}}_0  = \sum_{n=1}^{N}{w_t^{n}\bm{x}_{t}^{n}},
\end{equation}
where $\bm{x}_{t}^{n}$ indicates the intermediate result of the $n$-th channel at the current $t$-th step of the reverse diffusion process and $w_t^{n}$ denotes the weight of $n^{th}$ channel of time $t$. While enjoying simplicity, the estimation of $\boldsymbol{\hat{x}}_0$ in Eq. \eqref{eqIII6} admits several desirable properties as follows.\\

\textbf{Proposition:} The starting pure noise images, $\bm{x}_{T}^{n}$ with $n\in [1,\dots, N]$, can be considered as a set of denoised images from the same ground truth $\boldsymbol{{x}}_0$ by gradually adding a sequence of independent noise to $\boldsymbol{{x}}_0$, i.e., $\bm{x}_t = \bm{x}_{t-1} + \sqrt{\sigma_t^2 - \sigma_{t-1}^2} \bm{z}_{t-1}$,  $\bm{z}_t \mathop{\sim} \limits^{i.i.d.} \mathcal{N}(\bm{0}, \bm{I})$. Then $\boldsymbol{\hat{x}}_0 = \sum_{n=1}^{N}{w_k\bm{x}_{t}^{n}}$ is an unbiased estimate of $\boldsymbol{{x}}_0$ when $N\to +\infty$ and $w_t^{n} = \frac{1}{N}$.\\

\textbf{Remark:} The variance expectation of the proposed estimation $\boldsymbol{\hat{x}}_0 = \frac{1}{N}\sum_{n=1}^{N}{\bm{x}_{t}^{n}}$ is expected to be smaller than the variance expectation with single $\bm{x}_{t}^{n}$. Since $\mathbb{E}\left(\text{Cov}\left( \boldsymbol{z}_{t_1}^{n_1},\boldsymbol{z}_{t_2}^{n_2} \right)\right) =0,\ \forall t_1\ne t_2,\ n_1\ne n_2, \boldsymbol{z}_{t}^{n} \mathop{\sim} \limits^{i.i.d.} \mathcal{N}(\bm{0}, \bm{I})$, so $\mathbb{E}\left(\text{Var}\left( \frac{1}{N}\sum_{n=1}^N{\boldsymbol{x}_{t}^{n}} \right)\right) =\frac{1}{N}\mathbb{E}\left(\text{Var}\left( \boldsymbol{x}_{t}^{n} \right)\right)$. This means our proposed can improve the stability of reconstruction. Indeed, the proposed estimation $\boldsymbol{\hat{x}}_0 = \frac{1}{N}\sum_{n=1}^{N}{\bm{x}_{t}^{n}}$ admit a reliable solution and could effectively facilitate the enforcement of data consistency term during the reconstruction process.

Considering the estimation of  $\hat{x_0}$, we can establish the optimization model as following:
\begin{equation}\label{eqIII13}
\bm{x}_{t-1}=\min _{\bm{x}_{t}} \frac{1}{2}\left\|\bm{A} \sum_{n=1}^{N} w_{t}^{n} \bm{x}_{t}^{n}-y\right\|_{2}^{2}+\eta_{t} M\left(\sum_{n=1}^{N} w_{t}^{n} \bm{x}_{t}^{n}\right)
\end{equation}
where $w_t^n$ represents the weight of $n^{th}$ noisy images from time $t$ and satisfy with $\sum_{n=1}^{N}w_t^n=1$.

\subsection{Optimization Procedure}
Eq. \eqref{eqIII13} demonstrates the VE-SDE prior is posed on the $n$ images rather than a single image $\bm{x}_t^n$. As evident from our observations, Eq. \eqref{eqIII13} unmistakably brings to light the inherent nexus between data consistency and the diffusion generative model. A viable avenue for resolution emerges with the introduction of an additional fidelity term aimed at solidifying the alignment between these two pivotal terminologies. In this pursuit, we introduce $u_t^n$ as a substitution for $x_t^n$, which is treated as a bridge to connect $\hat{\bm{x}_0}$ and $M\left(\bm{x}_t\right)$. Actually, we can simply the Eq. \eqref{eqIII13} into a series of separable optimization problems. Consequently, the manifestation of Eq. \eqref{eqIII13} takes on a transformed disposition, evolving into an intricately formulated optimization challenge characterized by the subsequent expression:
\begin{equation}\label{eqIII14}
\begin{aligned}
\left\{\bm{x}_{t-1}, \bm{u}_{t-1}\right\}=&\min _{\left\{\bm{x}_{t}, \bm{u}_{t}\right\}} \frac{1}{2}\left\|\bm{A} \sum_{n=1}^{N} w_{t}^{n} \bm{x}_{t}^{n}-\bm{y}\right\|_{2}^{2} \\
&+\eta_{t} M\left(\sum_{n=1}^{N} w_{t}^{n} \bm{u}_{t}^{n}\right), \\
&\text { s.t. }, \bm{x}_{t}^{n}=\bm{u}_{t}^{n}, n=1, \ldots, N.
\end{aligned}
\end{equation}

Eq. \eqref{eqIII14} poses a problem constrained by the requirement $\bm{x}_t^n=\bm{u}_t^n$, and under certain circumstances, it can be transformed into an unconstrained optimization problem as below
\begin{equation}\label{eqIII15}
\begin{aligned}
\left\{\bm{x}_{t-1}, \bm{u}_{t-1}\right\}=&\min _{\left\{\bm{x}_{t}, \bm{u}_{t}\right\}} \frac{1}{2}\left\|\bm{A} \sum_{n=1}^{N} w_{t}^{n} \bm{x}_{t}^{n}-\bm{y}\right\|_{2}^{2} \\
&+\eta_{t} M\left(\sum_{n=1}^{N} w_{t}^{n} \bm{u}_{t}^{n}\right)+\frac{\beta_t}{2} \sum_{n=1}^{N}\left\| \bm{x}_{t}^{n}-\bm{u}_{t}^{n}\right\|_{2}^{2} ,
\end{aligned}
\end{equation}
where $\beta_t>0$ is a factor. It's evident from Eq. \eqref{eqIII15} that the inclusion of $\bm{u}_t$ serves a dual purpose. On one hand, it enforces the reconstructed outcomes to align reasonably well with the measurements; on the other hand, it imposes a constraint to confine the divergence between the reconstruction outcomes and their generative counterparts within specific bounds. Notably, the term $\frac{\beta_t}{2}||\bm{x}_t-\bm{u}_t||_2^2$ constitutes a vital component of data consistency. Consequently, the optimization of Eq. \eqref{eqIII14} can be viewed as a process bifurcated into two distinct steps:
\begin{equation}\label{eqIII16}
\begin{aligned}
\bm{x}_{t-1}=&\min _{\bm{x}_{t}} \frac{1}{2}\left\|\bm{A} \sum_{n=1}^{N} w_{t}^{n} \bm{x}_{t}^{n}-\bm{y}\right\|_{2}^{2}+\frac{\beta_t}{2} \sum_{n=1}^{N}\left\| \bm{x}_{t}^{n}-\bm{u}_{t}^{n}\right\|_{2}^{2} \\
\bm{u}_{t-1}=&\min _{\bm{u}_{t}} \frac{1}{2} \sum_{n=1}^{N}\left\| \bm{x}_{t}^{n}-\bm{u}_{t}^{n}\right\|_{2}^{2}+\eta_{t} M\left(\sum_{n=1}^{N} w_{t}^{n} \bm{u}_{t}^{n}\right).
\end{aligned}
\end{equation}
Regarding Eq. \eqref{eqIII16}, which introduces a new term to ensure data consistency. Several iterative methods are available for achieving the best solution. However, ensuring convergence is difficult while determining the appropriate iterative step length presents a practical hindrance to optimization efforts. The gradient-based descent technique is a popular solution avenue, commonly embraced in recent diffusion models \cite{chung2022improving} \cite{chung2022diffusion}. However, these approaches often demonstrate susceptibility to noise and perturbations, which is a common occurrence in real-world scenarios. Thus, we have chosen to employ an advanced optimization strategy for enhancing the performance of Eq. \eqref{eqIII16}. Note that the update of $\bm{x}_{t-1}$ comes from n images, we can treat $\bm{x}_{t-1}=\sum_{n=1}^{N}{w_t^n\bm{x}_t^n}$ and obtain:
\begin{equation}\label{eqIII17}
\begin{aligned}
\bm{x}_{t-1/2}=\min _{\bm{x}_{t}} \frac{1}{2}\left\|\bm{A} \sum_{n=1}^{N} w_{t}^{n} \bm{x}_{t}^{n}-y\right\|_{2}^{2}.
\end{aligned}
\end{equation}
By incorporating the wisdom of simultaneous iteration reconstruction technique \cite{trampert1990simultaneous}, in respect to the Eq. \eqref{eqIII10} step, we can update it with:
\begin{equation}\label{eqIII18}
\begin{aligned}
x_{t-\frac{1}{2}}^{n}=x_{t}^{n}+D A^{T} E\left(y-A \sum_{n=1}^{N} w_{t}^{n} x_{t}^{n}\right),\quad n=1,\ldots, N
\end{aligned}
\end{equation}
Here, $D$ and $E$ are diagonal matrices encompassing the reciprocal of the column and row sums of the system matrix. Specifically, $d_{jj}$ corresponds to one divided by the sum of elements in row $i$, i.e., $d_{jj}=1/{\sum_{i} h_{ij}}$, and similarly, $e_{jj}$ equals one divided by the sum of elements in row $j$, i.e., $e_{jj}=1/{\sum_{j} h_{ij}}$. These matrices serve to balance the impact of rays hitting individual pixels and pixels intersected by each ray. The transpose of the matrix, denoted as $\bm{A}^T$, facilitates the back-projection of projection images onto the reconstruction area. It delineates the pixels influenced by a particular ray's trajectory. This mechanism enables the derivation of the updated $\bm{x}_{t-1}^n$ from the previously updated $\bm{x}_{t-1/2}^n$, thus allowing for the subsequent progression of the process.  
\begin{equation}\label{eqIII19}
\begin{aligned}
\bm{x}_{t-1}^n=\bm{x}_{t-1/2}^n+\beta_t(\bm{x}_{t}^n-\bm{u}_{t}^n), \quad n=1, \ldots, N .
\end{aligned}
\end{equation}
If all the reconstruction procedure is finished, the estimated $\hat{\bm{x}_0}=\sum_{n=1}^{N} w_{0}^{n} \bm{x}_{0}^{n}$. Regarding the second step of Eq. \eqref{eqIII6}, it outlines the diffusion model's advancement process, utilizing the VE-SDE approach for updates. Initially, the generation procedure for the diffusion model is a generalized numerical solution derived from the inverse SDE. In contrast to generating samples using the numerical SDE solver, we opt for the Predictor-Corrector (PC) sampler \cite{songscore}. This choice is driven by its superior performance when dealing with VE-SDEs. In the context of PC samplers, the predictor entails a numerical solver for the reverse-time SDE, while the corrector encompasses any Markov Chain Monte Carlo (MCMC) technique reliant solely on the scores. As previously mentioned, the SDE update procedure is partitioned into predictor and corrector updates. Concerning the predictor update, we can perform enhancements to refine the estimation of the model parameters.
\begin{equation}\label{eqIII20}
\bm{u}_{t}^n= \bm{x}_t^n+\left(\sigma_{t}^{2}-\sigma_{t-1}^{2}\right) \bm{s}_{\theta^{*}}\left(\bm{x}_{t}^n, t\right)+\sqrt{\sigma_{t}^{2}-\sigma_{t-1}^{2}} \bm{z}, 
\end{equation}
where the parameter $\sigma_t>0$ embodies a monotonically rising function in relation to time $t$, as stipulated by \cite{songscore}. $z~N\left(0,1\right)$ is random noise that follows a Gaussian distribution, which is added by \cite{songscore} and can prevent the model from reaching the local optimization solution. Regarding the corrector step, the update adheres to the subsequent formulation:
\begin{equation}\label{eqIII21}
\bm{u}_{t}^n= \bm{u}_{t}^n+\epsilon_{t-1} \bm{s}_{\theta^{*}}\left(\bm{x}_{t}^n, t\right)+\sqrt{2\epsilon_{t-1}} \bm{z},
\end{equation}
where $\epsilon_{t-1}$ is the step size of at time $t-1$.
%%%%%%%%%%%%%
\subsection{Overall Clarification}

\begin{figure}[htp]
    \centering
    \vspace{-0.cm}
    \includegraphics[scale=0.54]{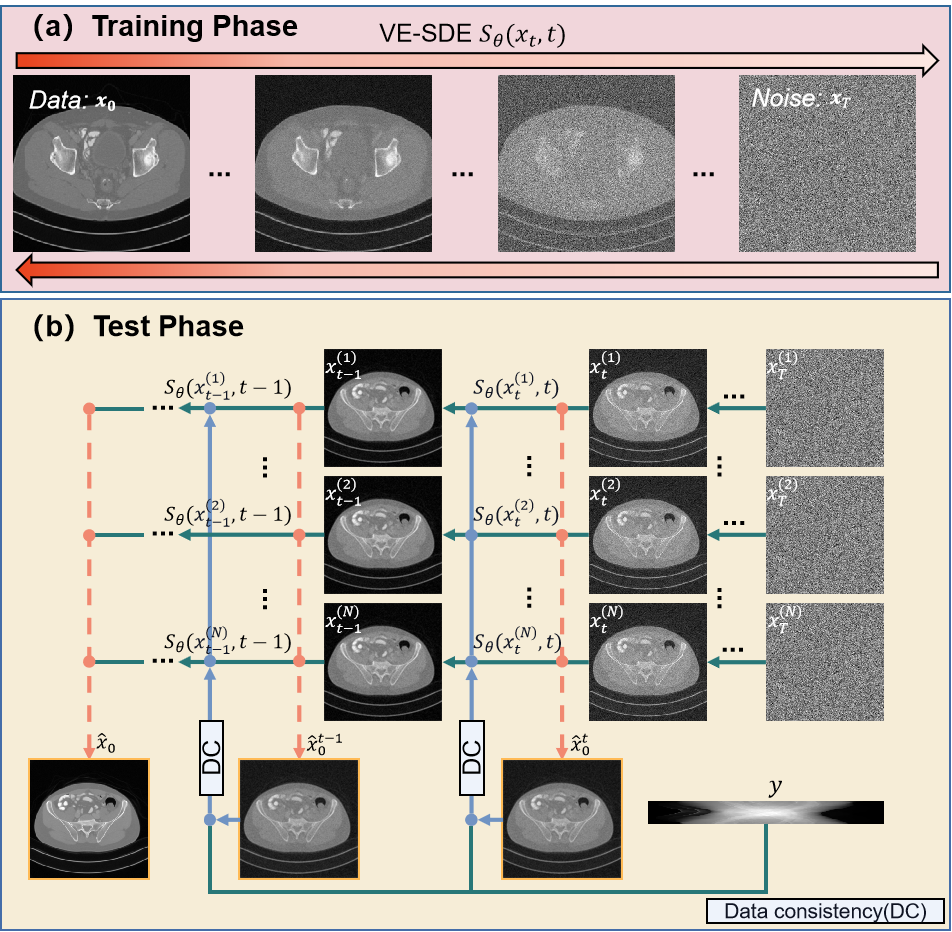}
    \caption{ Overall architecture of DOSM network. (a) and (b) represent the training and testing of the network, respectively. The green line represents the diffusion of the backward SDE, the blue line represents the constraint of the data consistency policy, and the orange line represents the estimated $\bm{\hat{x}}_0$.}
    \label{mainfig1}
\end{figure}

The DOSM algorithm's overarching structure is depicted in Fig. \ref{mainfig1}, leveraging the insights from the preceding analysis. We first train the VE-SDE model at the training phase. For the test phase, with a set of input noise images across $N$ channels, we harness both the data consistency principle and the learned VE-SDE $S_\theta$ to derive $\bm{x}_{T-1}^n$ for $n=1, \ldots, N$. This process sets the stage for computing $\hat{x}_0$ at the penultimate time instance, specifically denoted as $\hat{x}_0^{T-1}$. Importantly, this updated $\hat{x}_0^{T-1}$ becomes pivotal for enhancing data consistency at the preceding time step of $T-2$. The iterative updating methodology results in the acquisition of the ultimate estimation of $\hat{x}_0$.

\begin{figure}[t]
    \centering
    \vspace{-0.cm}
    \includegraphics[scale=0.48]{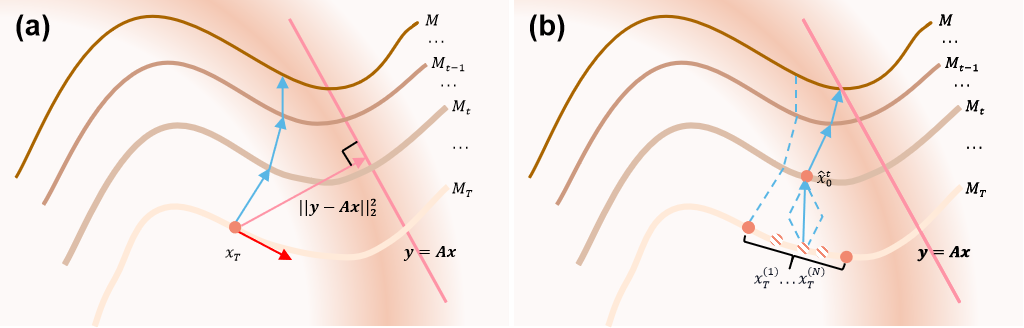}
    \caption{Advantages of the DOSM reconstruction step (b) over the traditional diffusion model step (a).
The equipotential surface is denoted as $M_t$, and the optimization process is centered around identifying the intersection solution between $M_t$ and the data consistency policy. The red and pink arrows indicate the directions of the diffusion model and the data consistency policy. The blue arrow indicates the direction in which the search optimization is advancing.}
    \label{manifold}
\end{figure}

In conventional data consistency approaches, the search direction remains unconstrained, lacking a definitive objective for computing a dependable gradient, which in turn contributes to the instability of the optimization process. It is the reason why the other SGM-based has a relative poor convergence, as the MCG and DPS in Fig. \ref{motion}. This instability ultimately hampers the accurate retrieval of the actual $\bm{x}_0$, as shown in Fig. \ref{manifold}(a). In contrast, our proposed DOSM fully addresses the challenges in Fig. \ref{manifold}(b). Firstly, it establishes a stable target, denoted as $\hat{x}_0$ and validated by \textbf{Proposition}, right from the outset of the process. This grounded objective provides a clear optimization direction, lending stability to the overall procedure. Secondly, owing to the presence of $N$-channel noisy images at time $t$, the DOSM inherently encompasses $N$ in distinct directions. This amalgamation of directions not only enriches the search landscape but also contributes significantly to fortifying the overall stability of the reconstruction model. Collectively, these factors substantiate a better robust convergence of our DOSM, as illustrated in Fig. \ref{motion}.

%\text{And,\ }Var\left( \boldsymbol{x}_t \right) =nVar\left( \frac{1}{n}\sum_{i=1}^n{\boldsymbol{x}_{t}^{i}} \right) $

\begin{figure*}[htp]
    \centering
    \vspace{-0.cm}
    \includegraphics[scale=0.5]{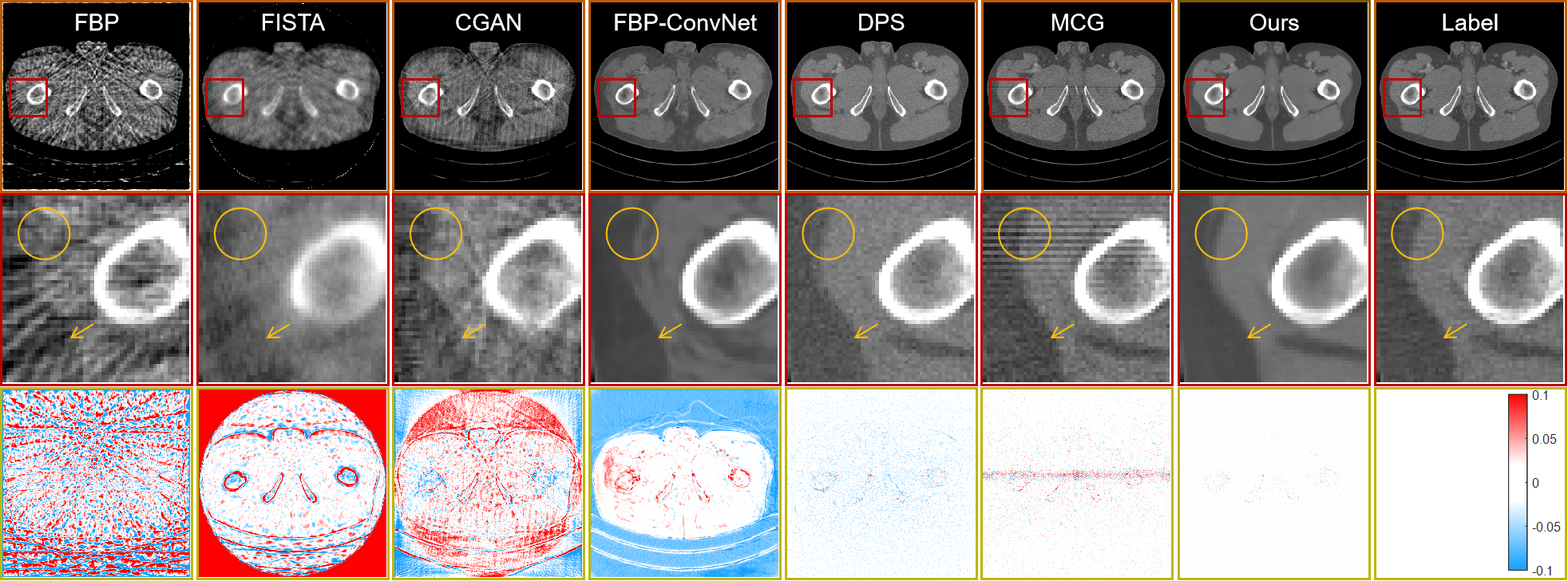}
    \caption{Representative reconstructed results of simulated AAPM CT from 23 views by different methods.  From top to bottom:  reconstructed results, region of interest (ROI), and difference images. The columns from left to right are FBP, FISTA, DPS, MCG, Ours and Label.}
    \label{23-viewAAPM}
\end{figure*} 
%%%%%%%%%%%%%%%%
\begin{figure*}[htp]
    \centering
    \vspace{-0.cm}
    \includegraphics[scale=0.5]{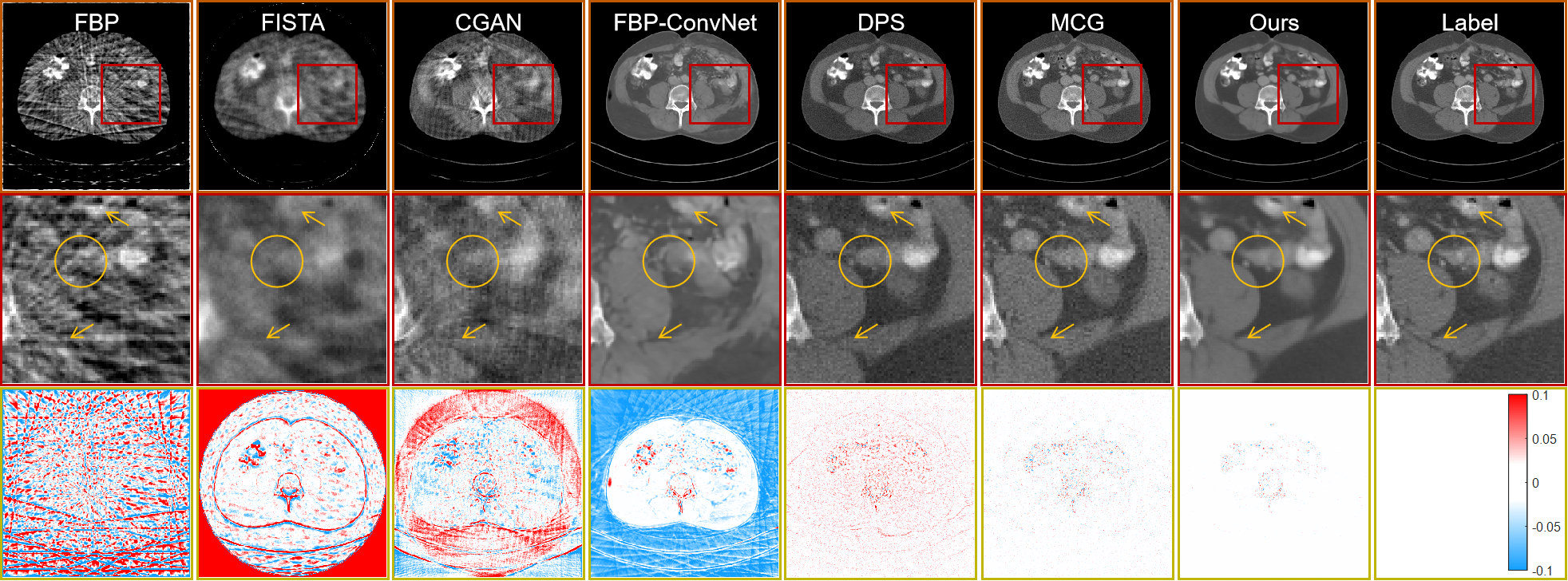}
    \caption{Similar to the illustration in Fig. \ref{23-viewAAPM} but from an alternate representative slice.}
    \label{23-viewAAPM-200}
\end{figure*} 

\begin{figure*}[htp]
    \centering
    \vspace{-0.cm}
    \includegraphics[scale=0.5]{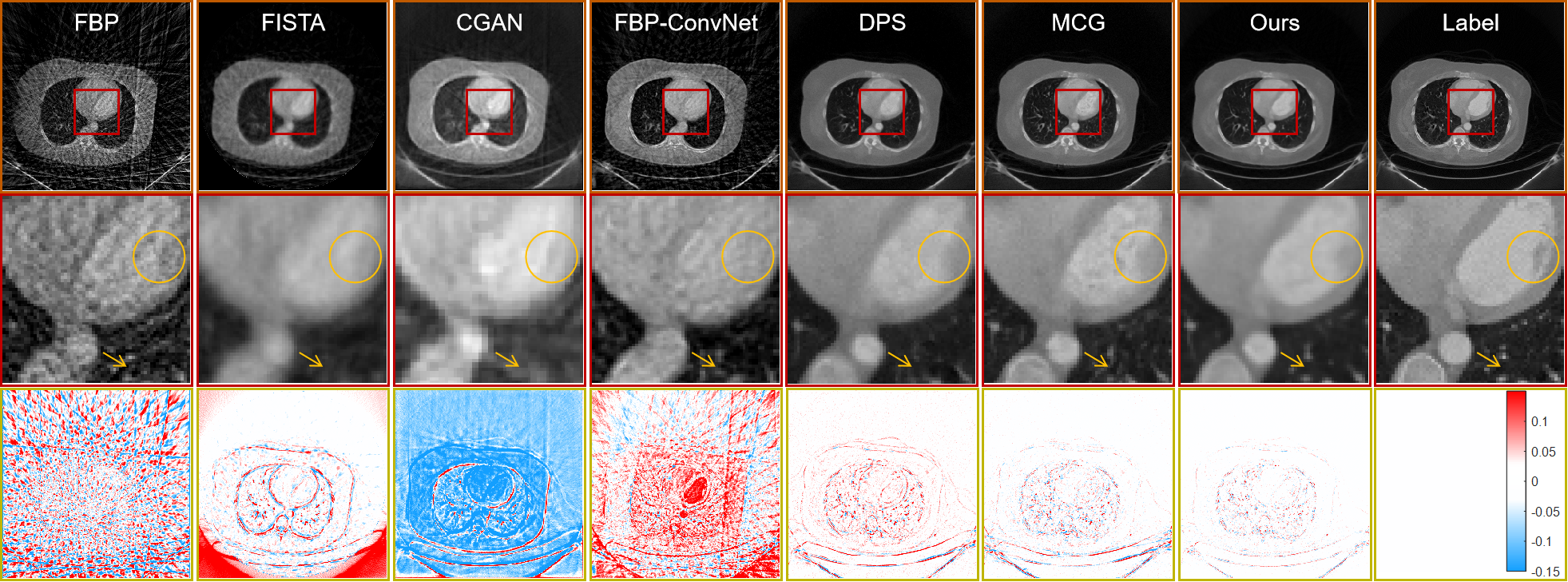}
    \caption{Representative reconstructed results of clinical cardiae datasets from 23 views by different methods. From top to bottom: reconstructed results, region of interest, and different images. The columns from left to right:  FBP, FISTA, DPS, MCG, Ours, and Label.}
    \label{23-viewGE}
\end{figure*} 

\begin{figure}[htp]
    \centering
    \vspace{-0.cm}
    \includegraphics[scale=0.35]{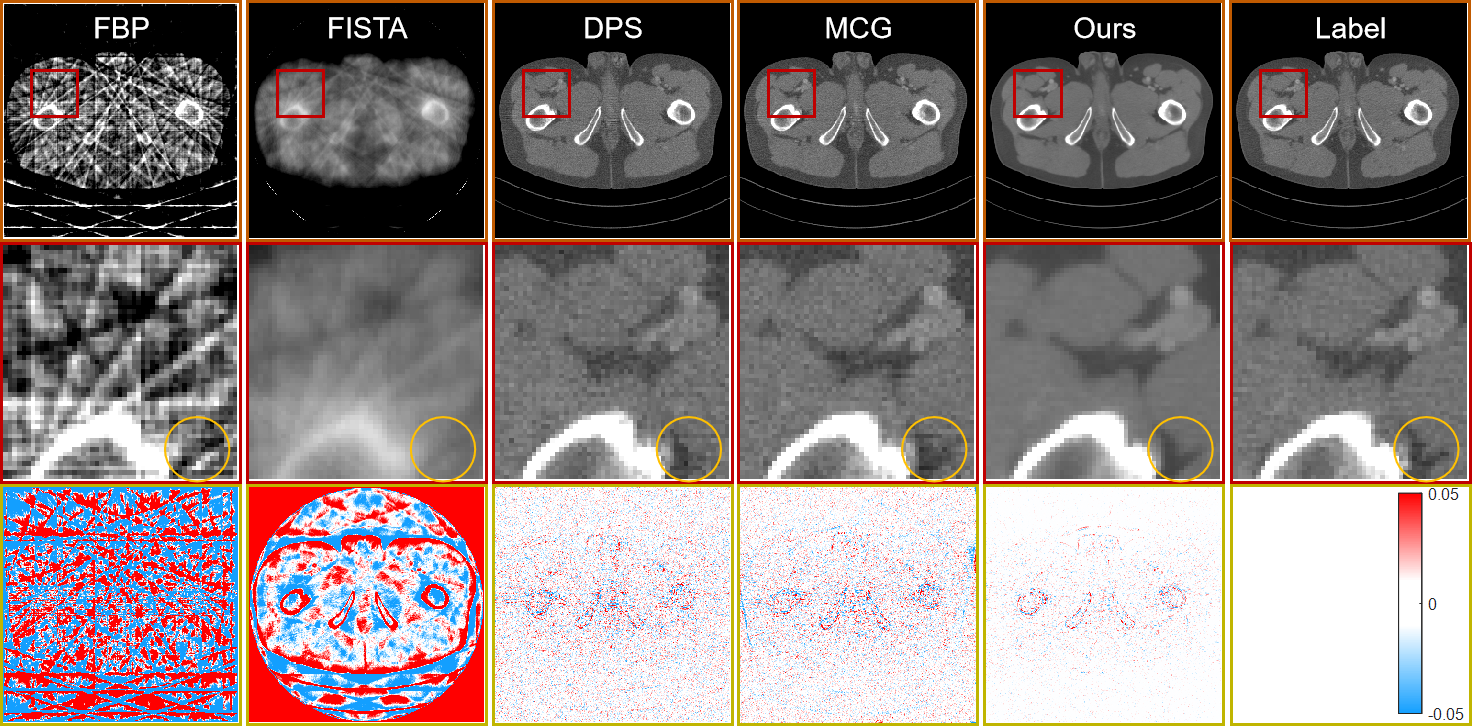}
    \caption{Representative reconstructed results of simulated AAPM CT from 10 views by different methods. From top to bottom: reconstructed results, ROI, and difference images.}
    \label{10-viewAAPM}
\end{figure} 

\begin{figure}[htp]
    \centering
    \vspace{-0.cm}
    \includegraphics[scale=0.35]{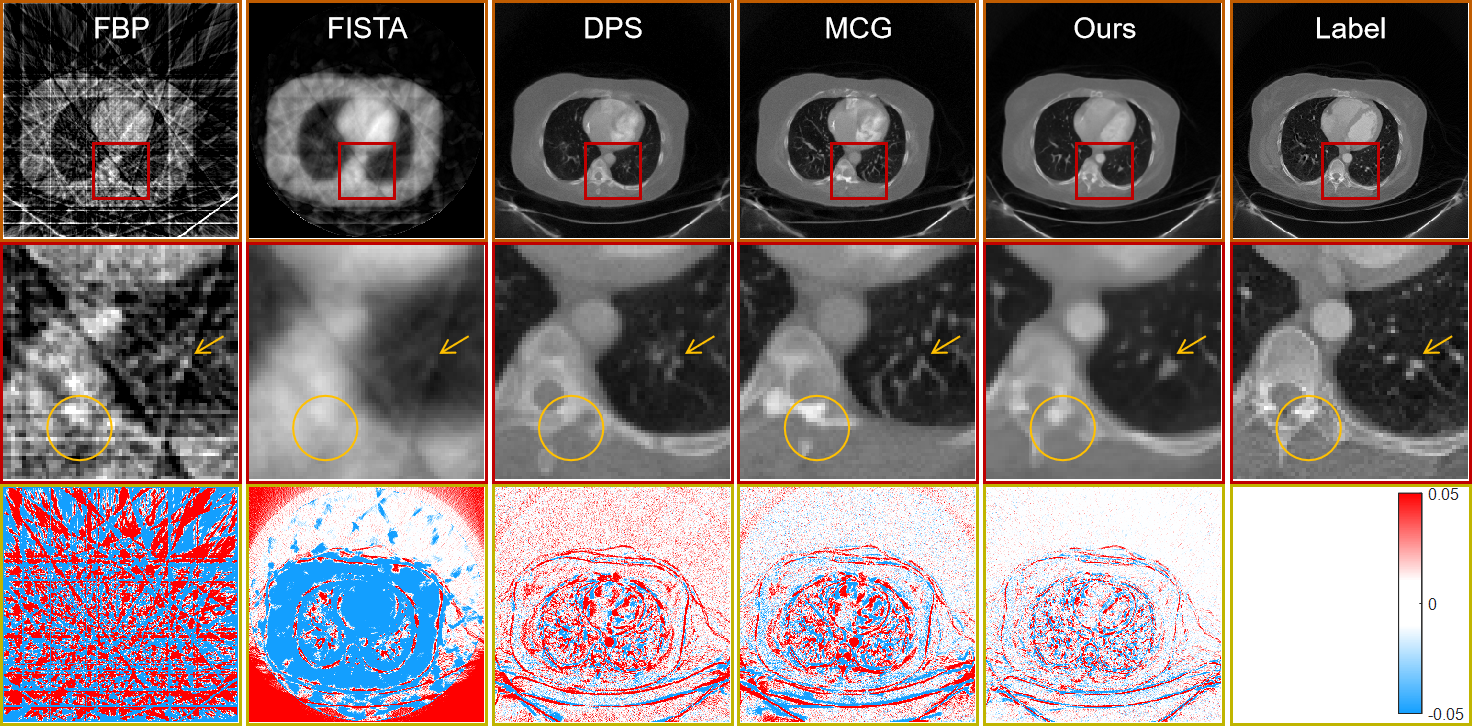}
    \caption{Representative reconstructed results of clinical cardiae datasets from 10 views by different methods. From top to bottom: reconstructed results, ROI, and difference image.}
    \label{10-view-GE}
\end{figure} 

%

%\begin{figure}[t]
    %\centering
    %\vspace{-0.cm}
    %\includegraphics[scale=0.3]{picture/iteration.png}
    %width=20cm,height=14cm
    %\caption{The process of Our-CGLS iterative convergence in a case is shown. We conducted four separate replicates of the experiment and found significant diversity in the results generated by the score-based Model under the same scoring function }
    %\label{iteration}
%\end{figure} 

\begin{figure}[t]
    \centering
    \vspace{-0.cm}
    \includegraphics[scale=0.3]{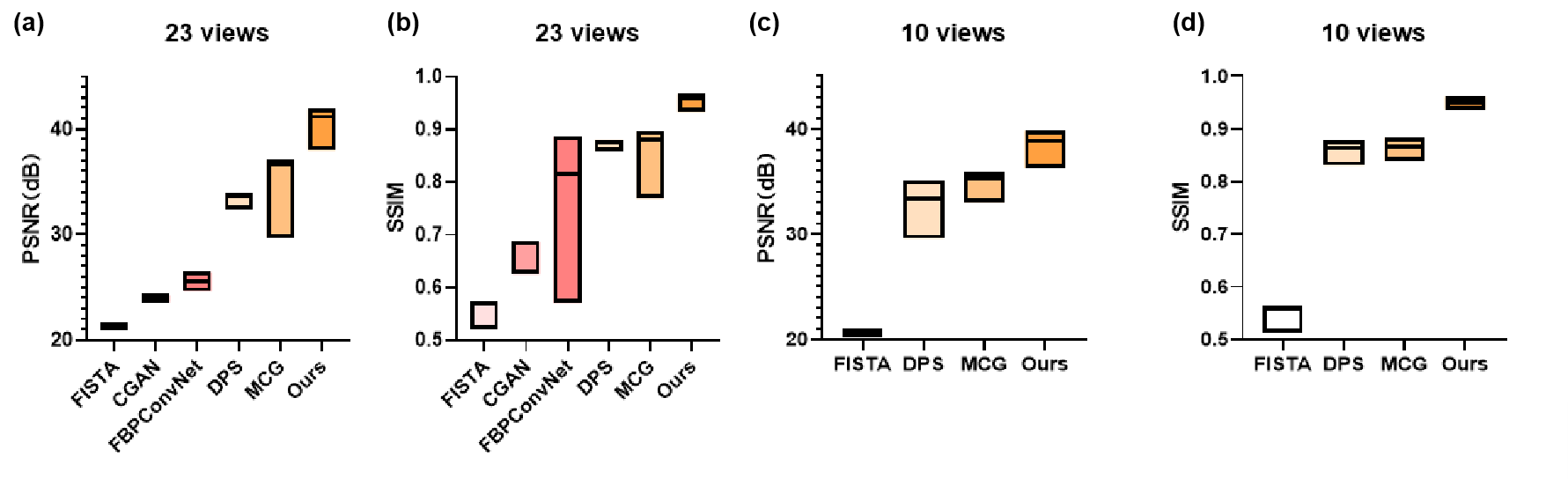}
    \caption{Visualized PSNR and SSIM quantitative results of AAPM CT datasets in 23 and 10 views.}
    \label{Phase}
\end{figure} 

% Please add the following required packages to your document preamble:
% \usepackage[table,xcdraw]{xcolor}
% If you use beamer only pass "xcolor=table" option, i.e. \documentclass[xcolor=table]{beamer}
\begin{table}[]
\centering
\caption{The statistics results in terms of PSNR and SSIM from 23 and 10 views in clinical cardiac datasets}
\label{tab:my-table}
\begin{tabular}{c|cccc}
\hline
{\color[HTML]{000000} } &
  \multicolumn{4}{c}{{\color[HTML]{000000} Clinical Cardiae Datasets}} \\ \cline{2-5} 
Method &
  \multicolumn{2}{c|}{23 Views} &
  \multicolumn{2}{l}{10 Views} \\ \cline{2-5} 
\textbf{} &
  \multicolumn{1}{c|}{PSNR} &
  \multicolumn{1}{c|}{SSIM} &
  \multicolumn{1}{l|}{PSNR} &
  \multicolumn{1}{l}{SSIM} \\ \hline
{\color[HTML]{000000} FBP} &
  \multicolumn{1}{c|}{{\color[HTML]{000000} 20.30}} &
  \multicolumn{1}{c|}{{\color[HTML]{000000} 0.374}} &
  \multicolumn{1}{c|}{15.16} &
  0.209 \\
{\color[HTML]{000000} FISTA} &
  \multicolumn{1}{c|}{{\color[HTML]{000000} 22.29}} &
  \multicolumn{1}{c|}{{\color[HTML]{000000} 0.618}} &
  \multicolumn{1}{c|}{21.84} &
  0.508 \\
CGAN &
  \multicolumn{1}{c|}{22.57} &
  \multicolumn{1}{c|}{0.623} &
  \multicolumn{1}{c|}{/} &
  / \\
FBP-ConvNet &
  \multicolumn{1}{c|}{24.31} &
  \multicolumn{1}{c|}{0.721} &
  \multicolumn{1}{c|}{/} &
  / \\
DPS &
  \multicolumn{1}{c|}{31.27} &
  \multicolumn{1}{c|}{0.763} &
  \multicolumn{1}{c|}{28.25} &
  0.657 \\
MCG &
  \multicolumn{1}{c|}{32.22} &
  \multicolumn{1}{c|}{0.774} &
  \multicolumn{1}{c|}{26.44} &
  0.699 \\
Ours &
  \multicolumn{1}{c|}{{\color[HTML]{FE0000} 33.93}} &
  \multicolumn{1}{c|}{{\color[HTML]{FE0000} 0.881}} &
  \multicolumn{1}{c|}{{\color[HTML]{FE0000} 32.11}} &
  {\color[HTML]{FE0000} 0.874} \\ \hline
\end{tabular}
\end{table}
%%%%%%

\section{Experiments}

In this study, we conduct our experimental investigations by harnessing the Python programming language within the PyTorch framework. All experiments are implemented on a high-performance computing system equipped with an NVIDIA RTX A6000 48GB graphics processing unit. The training procedure of VE-SDE follows with the guidelines recommended by Song et al\cite{songsolving}. Here, the Adam optimization algorithm is used and the learning rate is set to $2\times10^{-4}$. In configuring the noise variance, we establish fixed values of $\sigma_{\text{min}}=0.01$ and $\sigma_{\text{max}}=378$. The number of iterations within the sampling process is set as $2000$. Furthermore, a series of comparison methods are chosen, including FISTA \cite{beck2009fast}, CGAN \cite{ghani2018deep}, FBP-ConvNet \cite{jin2017deep}, DPS \cite{chung2022diffusion} and MCG \cite{chung2022improving}. To quantitatively assess and contrast the efficacy of the reconstruction outcomes, we employ two metrics: the Peak Signal-to-Noise Ratio (PSNR) and the Structural Similarity (SSIM). Elevated values of PSNR and SSIM correspond to heightened levels of reconstruction quality. Furthermore, our code and datasets are provided for review with the link.

\subsection{Data Description}
We first train and test our model using a simulated dataset coming from the AAPM 2016 CT Low Dose Challenge \cite{timmurphy.org}. For the simulation study, the clinical data from nine patients'  are used for training (4,000 slices), and the rest of the patient's data is used for testing (400 slices). 
To generate the 23 views and 10 views data, an equiangular fan-beam projection geometry is employed. We systematically generated data from both 23 and 10 views to facilitate the sparse-view CT reconstruction process. Regarding the scanning geometry configuration, the distances from the rotation center to the source and detector are 1500mm and 500 mm. The detector width is 41.3 cm with 720 detector elements.

The real cardiac clinical datasets used in ~\cite{xu2012low} is utilized in this experiment. The curved cylindrical detector array encompassed a total of 880 individual units, while the complete scan is composed of 2200 views. The field-of-view (FOV) diameter covers an expansive 49.8 × 49.8 $cm^2$, and the image matrix featured dimensions of 512 × 512 pixels. Notably, the distance from the X-ray source to the center and detector are set as 53.85 $cm$ and 103.68 $cm$. To validate the universality and exceptional performance of our approach, the model trained using simulation data is seamlessly applied to the testing of clinical cardiac data.
%%%
\subsection{Experimental results}
\subsubsection{Simulate Dataset Results} We first compare the reconstruction results with 23 views on the simulated clinical datasets, as shown in Figs. \ref{23-viewAAPM} and \ref{23-viewAAPM-200}. Upon scrutiny, it becomes evident that the FBP results exhibit an excessive proliferation of artifacts that obscure image intricacies, rendering it challenging to extract informative content beyond the skeletal structure. FISTA, as a classical iterative reconstruction technique, manages to enhance image quality compared to FBP, yet elusive finer details still persist. FBP-ConvNet, an exemplar of supervised deep learning approaches, brings about substantial improvements in image quality, accentuating large structures and edges. However, a notable drawback arises in the form of inaccurately predicted reconstruction details, a consequence of the profound information loss due to the ultra-sparse-view measurements.

Incorporating data consistency into the generation process, the score-based model significantly enhances the efficacy of image reconstruction. In the sparse-view CT reconstruction landscape, DPS \cite{chung2022diffusion} and MCG \cite{chung2022improving} stand as prevailing benchmarks. In comparison to traditional and typical deep learning-based methods, the score-based methodology consistently attains superior results, yielding heightened image quality. A closer examination within the region of interest (ROIs) reveals that the image edges delineated by yellow circles in the DPS result become blurred. Similarly, the MCG outcomes exhibit conspicuous artifacts and noise, particularly within the extracted ROI of Fig. \ref{23-viewAAPM}. This artifact's presence emanates from the intrinsic integration of estimated projection within MCG, causing an overly constrained data consistency enforcement during iterations. Additionally, image edges denoted by yellow arrows are also subjected to blurring.

In contradistinction to these contenders, our proposed approach showcases the most remarkable image quality, accompanied by a discernible preservation of image features and details. This assertion is fortified by the discernible clarity of the yellow circles and arrows within the ROIs. The difference map from our method demonstrates the closest proximity to zero, unequivocally signifying the supremacy of our approach over other baseline methods. Moreover, to demonstrate the out-performance of our method, the profiles along the yellow location with all reconstruction algorithms are shown in Fig. \ref{pixel-23-AAPM}. One can see that the profile reconstructed by our method is closer to the ground truth.

Likewise, we proceed to assess the efficacy of our methodology involving only 10 views for image reconstruction, as presented in Fig. \ref{10-viewAAPM}. The DPS and MCG baselines have a pronounced presence of noise within the reconstruction outcomes under such ultra-sparse-view condition, which is further confirmed by the extracted ROI. That demonstrates the data consistency strategies within these two methods struggle to adequately guide the generation process of the score-based model under ultra-sparse-view settings. Within the ROI featured in Fig. \ref{10-viewAAPM}, the textural intricacies captured within the yellow circles, as reconstructed by our method, significantly outperform those achieved by the alternative approaches. The difference map can also clearly demonstrate the gap between the reconstructed image and the ground truth obtained by our method is the smallest. The reconstruction's quantitative results from both 23 and 10 views in Fig. \ref{Phase} consistently demonstrate excellent reconstruction performance with the best PSNRs and SSIMs of our method.

\subsubsection{Cardiac Clinical Dataset Results} To further evaluate the efficacy and superior performance of our approach, we perform experimental validation on real clinical cardiac CT data. We employ the trained VE-SDE in above simulated clinical datasets for this real-world study. The reconstruction results from 23 and 10 views are showcased in Figs. \ref{23-viewGE} and \ref{10-view-GE}, respectively. It becomes apparent that the FBP results are marred by an excessive proliferation of artifacts, making it challenging to distinguish beneficial structures and details from these artifacts. While FISTA improves image quality in comparison to FBP, it falls short in capturing finer details. Conversely, FBP-ConvNet excels in enhancing image quality by highlighting prominent structures and edges. However, inaccuracies arise due to the extreme paucity of view measurements. This point is illustrated by the yellow circles in Fig. \ref{23-viewGE}.

Consistently outperforming traditional deep learning-based reconstructions, the score-based methodology consistently delivers elevated image quality. Upon closer examination of the specific ROI in Fig. \ref{23-viewGE}, it's evident that DPS results exhibit blurring at edges denoted by yellow circles. Furthermore, it introduces artifacts marked by yellow arrows. Similarly, the outcomes of MCG display conspicuous artifacts and noise, particularly within the extracted ROIs of Figs. \ref{23-viewGE} and \ref{10-view-GE}. While MCG introduces details and features, they are erroneous structures and artifacts that ultimately degrade image quality. Additionally, the image edges indicated by yellow arrows in Fig. \ref{23-viewGE} are inaccurate in the MCG results. Notably, the image details and edges introduced by MCG in Fig. \ref{10-view-GE} are also inaccurate. The discrepancy map in Figs. \ref{23-viewGE} and \ref{10-view-GE} vividly illustrate that the disparity between the reconstructed image and the ground truth obtained by our method is minimal.

\begin{figure}[t]
    \centering
    \vspace{-0.cm}
    \includegraphics[scale=0.3]{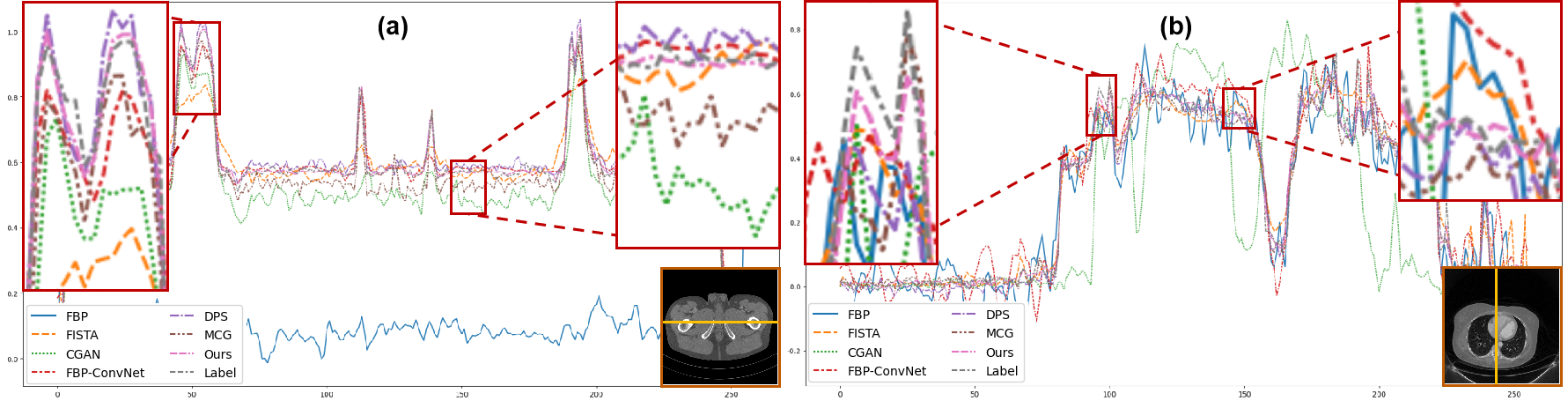}
    \caption{The intensity profiles along the specified yellow line in the reconstructed images. (a) represent 23 views of simulated dataset results. (b) represent 23 views of clinical cardiae datasets results.}
    \label{pixel-23-AAPM}
\end{figure}

Moreover, profile analyses along the yellow location with all reconstruction algorithms in Fig. \ref{pixel-23-AAPM} reveal that the profile reconstructed by our method closely approximates the ground truth. The quantitative reconstruction results from both 23 and 10 views, presented in Tables I, consistently underscore the exceptional reconstruction performance of our method, evidenced by superior PSNRs and SSIMs.

%%%%%%%%%%%%%%%%%%
\subsection{Ablation experiment}
In this section, ablation studies are performed to probe the effectiveness of the various modules of the proposed method. These experiments are carried out on the 23 views reconstruction task of simulated clinical data. We randomly selected 100 test data for ablation experiments.
\begin{figure}[htp]
    \centering
    \vspace{-0.cm}
    \includegraphics[scale=0.27]{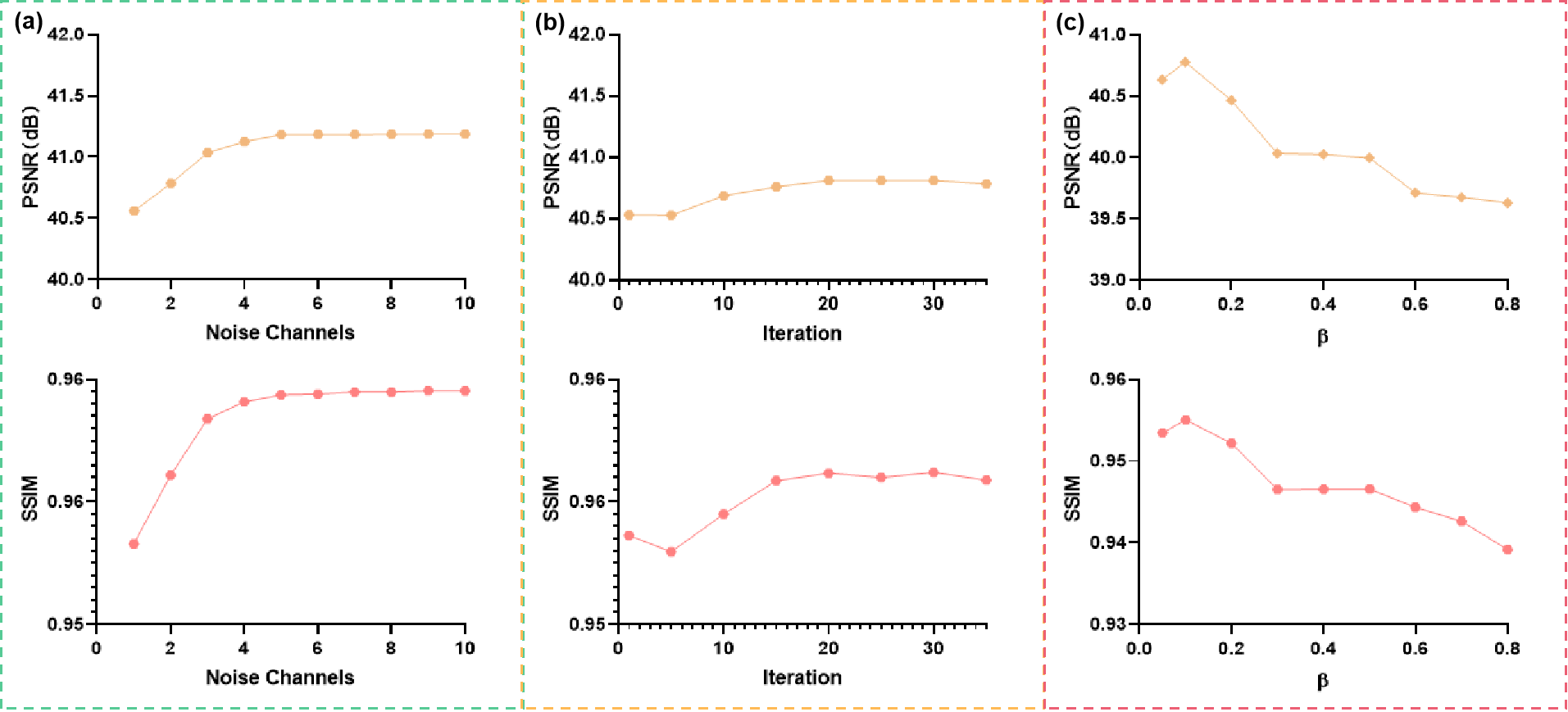}
    \caption{Comparison of the DOSM reconstruction performance under different experimental factors. We perform factor optimization using PSNR and SSIM. (a), (b), and (c) respectively represent the selection of parameters N for estimating $\bm{\hat{x}}_0$, the number of iterations, and the parameters of $\beta$.}
    \label{ablation}
\end{figure} 
%%%%%%%%%%%%%
\subsubsection{Selection of $N$}
The impact of $N$ on estimating $\bm{\hat{x}}_0$ is crucial. As the number of N increases, the constructed $\bm{\hat{x}}_0$ estimator can achieve better results. Figure \ref{ablation}(a) demonstrates that larger values of $N$ yield better quantitative outcomes, while an N of 5 maintains stable performance. However, higher $N$ increases computational costs. Our study finds a balance between performance and efficiency by choosing $N$ as 5.
\subsubsection{Iteration number setting within data consistency}
To provide effective guidance for the data consistency strategy during the reconstruction process, the process underwent multiple iterations to ensure the optimal alignment with data consistency requirements. The quantitative evaluation results of our approach are shown in Fig. \ref{ablation}(b), illustrating the impact of different iteration settings. The results indicate that as the iteration number increases, the reconstruction outcomes are enhanced. It's worth noting that the performance is stable with the data consistency settings when the iteration is set as 20 iterations. The selection of 20 iterations  within our data consistency strategy achieves a balance between reconstruction performance and computational efficiency.
\subsubsection{Selection of $\beta$}
As shown in Eq. \eqref{eqIII15}, the new data consistency strategy introduces an additional term to harmonize the impact of the current noisy data and the reconstruction outcomes generated through VE-SDE. Here, all $\beta_t$ are set as the same $\beta$. We delved further into the influence of $\beta$, as illustrated in Fig. \ref{ablation}(c). Evidently, an optimal factor can enhance the quantitative outcomes. In our investigation, this factor is designated as $\beta=0.1$.
\begin{figure}[t]
    \centering
    \vspace{-0.cm}
    \includegraphics[scale=0.5]{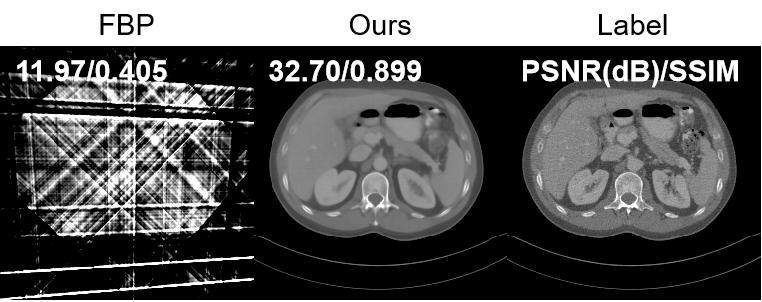}
    \caption{4 views of the reconstruction. results }
    \label{4-viewAAPM}
\end{figure} 
%%%
\section{Discussions and Conclusion}
While our proposed methods have demonstrated superior performance compared to other comparative techniques, it is imperative to address certain subtleties that demand attention. First, we acknowledge that our method surpasses classical score-based models in terms of computational time cost. This divergence emerges from the essential need to estimate $\bm{x}_0$ for establishing the data consistency. It's worth noting that this estimation process can be parallelized, effectively mitigating the computational overhead in practical implementation. Our proposed method has exhibited strong performance across both 23 and 10 views. However, a pertinent question arises: Can this prowess extend to even fewer views, such as 4 views. As illustrated in Fig. \ref{4-viewAAPM}, our approach continues to successfully reconstruct a good image even with only 4 views, while refining the image details remains a challenge. The pursuit of more advanced data consistency strategies and sophisticated score-based generative models emerges as an intriguing avenue for 4 views of CT reconstruction. In this study, we underscored the significance of achieving a higher-quality estimation of $\bm{x}_0$ holds significance. The potential to utilize a prior image as an improved $\bm{x}_0$ for guidance in reconstruction holds promise. 

%A viable approach entails employing a score-based generative model to reconstruct an initial image, subsequently utilizing it as a prior for guiding successive reconstruction endeavors.
%
 The proposed general framework exhibits remarkable reconstruction prowess for ultra-sparse-view CT reconstruction. Key attributes can be succinctly outlined as follows. Firstly, leveraging insights from conventional iterative reconstruction methods enhances reconstruction stability and elevates image quality. Moreover, an augmented data consistency paradigm is developed, seamlessly integrating $\bm{x}_0$ estimation. This two-fold approach significantly bolsters model robustness, where improved $p(\bm{x}_0)$ directly correlates with heightened image quality. Lastly, we introduce a discrepancy term to align the original data-driven prior with the SGM counterpart. Through extensive experiments on diverse datasets, including simulations and real clinical scenarios, our approach consistently demonstrates superior performance and unwavering stability.
%\section{Acknowledgments}
%We would like to express our sincere thanks to Jianjia Zhang and Zirong Li from the School of Biomedical Engineering of Sun Yat-sen University for polishing the expression of research.
\section{Acknowledgments}
We would like to express our sincere thanks to Jianjia Zhang and Zirong Li from the School of Biomedical Engineering of Sun Yat-sen University for polishing the expression of research.
%\section*{Acknowledgment}
\bibliographystyle{IEEEtran}
\bibliography{main}

\end{document}